%% file: acl2023.tex
\pdfoutput=1

\documentclass[11pt]{article}

\usepackage{ACL2023}

\usepackage{times}
\usepackage{latexsym}

\usepackage[T1]{fontenc}

\usepackage[utf8]{inputenc}

\usepackage{microtype}

\usepackage{inconsolata}

\usepackage{graphicx}
\usepackage{bm}
\usepackage{mathrsfs}
\usepackage{algorithm}
\usepackage{algorithmic}
\usepackage{comment}
\usepackage{multirow}
\usepackage{amsmath}
\usepackage{booktabs}
\usepackage{amssymb}
\usepackage{color}
\usepackage{cleveref}
\crefname{section}{§}{§§}
\input{Paragraphs/definition}
\usepackage{listings}
\newfloat{listing}{tb}{lst}{}
\floatname{listing}{Listing}

\title{\includegraphics[scale=0.03]{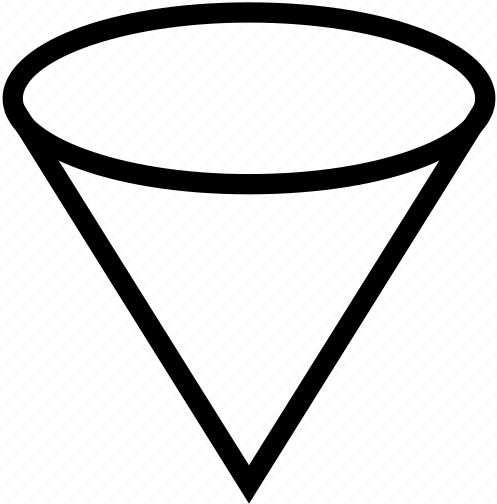}CONE: An Efficient COarse-to-fiNE Alignment Framework for \\ Long Video Temporal Grounding}
\author{Zhijian Hou$^{1}$\thanks{\ \ \ Indicates equal contribution. This work was done during the first and second authors' internships in MSR Asia.}, Wanjun Zhong$^{2*}$, Lei Ji$^3$, Difei Gao$^4$, Kun Yan$^5$, \\ \textbf{Wing-Kwong Chan$^1$, Chong-Wah Ngo$^6$, Mike Zheng Shou$^4$ and Nan Duan}$^3$ \\
    $^1$ City University of Hong Kong \quad
    $^2$ Sun Yat-sen University  \\
    $^3$ Microsoft Research Asia \quad 
    $^4$ National University of Singapore \\
    $^5$ Beihang University \quad 
    $^6$ Singapore Management University  \\
    {\tt \small \{zjhou3-c@my.,wkchan@\}cityu.edu.hk; 
     \tt \small zhongwj25@mail2.sysu.edu.cn; 
     \tt \small \{leiji,nanduan\}@microsoft.com} \\
    { \tt \small difei.gao@vipl.ict.ac.cn; 
      \tt \small kunyan@buaa.edu.cn;  
      \tt \small cwngo@smu.edu.sg;
      \tt \small mike.zheng.shou@gmail.com}\\
}

\begin{document}
\maketitle
\begin{abstract}

This paper tackles an emerging and challenging problem of long video temporal grounding~(VTG) that localizes video moments related to a natural language (NL) query.
Compared with short videos, long videos are also highly-demanded but less explored, which brings new challenges in   higher inference computation cost and weaker multi-modal alignment.
To address these challenges, we propose \modelname, an efficient  COarse-to-fiNE alignment framework. 
\modelname~is a plug-and-play framework on top of existing VTG models to handle long videos through a sliding window mechanism.
Specifically, CONE (1) introduces a query-guided window selection strategy to speed up inference, and (2) proposes a coarse-to-fine mechanism via a novel incorporation of contrastive learning to enhance multi-modal alignment for long videos.
Extensive experiments on two large-scale long VTG benchmarks consistently show both substantial performance gains (e.g., 3.13\%$\xrightarrow{\text{+119\%}}$6.87\% on MAD) and state-of-the-art results. 
Analyses also reveal higher efficiency as the query-guided window selection mechanism accelerates inference time by 2x on Ego4D-NLQ and 15x on MAD while keeping SOTA results. Codes have been released at \url{https://github.com/houzhijian/CONE}. 
\end{abstract}

\input{Paragraphs/1-Intro_new}
\input{Paragraphs/5-RelatedWork}

\input{Paragraphs/2-task}

\input{Paragraphs/3-approach}

\input{Paragraphs/4-experiment}

\input{Paragraphs/6-conclusion}
\input{Paragraphs/limitation.tex}

\section*{Ethics Statement}
The present study was conducted in accordance with ethical principles.
This study involved the analysis using publicly available data and did not involve any human participants, and potential risks about credentials or privacy. Therefore, no ethical clearance was required and there were no potential risks associated with the conduct of this research.

\section*{Acknowledgements}
We thank the anonymous reviewers for their insightful feedback. This research was partially supported by CityU MF\_EXT project number 9678180.


\bibliography{anthology,custom}
\bibliographystyle{acl_natbib}

\appendix
\label{sec:appendix}
\input{Paragraphs/appendix.tex}

\end{document}


\maketitle

\appendix
\section{Model Details}
\subsection{Experiment implementation Details}
Parameter optimization is performed by
AdamX. We set the learning rate to 1e-4 for Moment-DETR and 1e-5 for the visual adapter, respectively. We set the batch size to 32 and adopt the early stopping strategy. Finally, we use Non-Maximum Suppression~(NMS) with a threshold of 0.5 as post-processing.

\paragraph{Moment-DETR backbone.} The experiments are conducted on one P100 GPU.
For the transformer module inside Moment-DETR, we set the hidden size $d$ to 256, the layer number in encoder/decoder to 2, and the moment query number to 5. 
We set the window length to 90 video features~(48 seconds) for Ego4D-NLQ and 125 video features~(25 seconds) for MAD dataset. 
We train Ego4d for 150 epochs and MAD for 30 epochs, and the training time is about 3 hours for Ego4d and 24 hours for MAD dataset. 
We set the pre-filtering window number to 10 for Ego4d and 30 for the MAD dataset during inference. 

\paragraph{2D-TAN backbone.} The experiments are conducted on two V100 GPUs.
We set the window length to 64 video features~(34.1 seconds) for Ego4D-NLQ and 128 video features~(25.6 seconds) for MAD dataset. 
We train Ego4D for 150 epochs and MAD for 5 epochs, and the training time is about 3 hours for Ego4d and 24 hours for MAD dataset. 
We set the pre-filtering window number to 5 and 15 for the Ego4D and MAD datasets respectively during inference.

\subsection{Training Loss Details}


\bibliography{anthology,custom}
\bibliographystyle{acl_natbib}

%% file: Paragraphs/definition.tex
\newcommand{\modelname}{{\usefont{T1}{ptm}{m}{n}CONE}}
\newcommand{\wanjun}[1]{{[\textcolor{blue}{wanjun: #1}]}}
\newcommand{\zhijian}[1]{{[\color{cyan}{zhijian: #1}]}}
\newcommand{\lj}[1]{{\color{cyan}[lj: #1]}}
\newcommand{\zs}[1]{{\color{blue}[Shou: #1]}}

%% file: Paragraphs/1-Intro_new.tex
\section{Introduction}
Video temporal grounding~\cite{anne2017localizing,gao2017tall} aims to locate specific moments in an untrimmed video relevant to a textual user query.
This is a crucial task in multi-modal video understanding and can be applied to many practical applications such as video retrieval~\cite{xu2016msr}, video editing, and video question answering~\cite{lei-etal-2018-tvqa,lei-etal-2020-tvqa}. 

\begin{figure}[t]
\begin{center}
\includegraphics[width=0.99\linewidth]{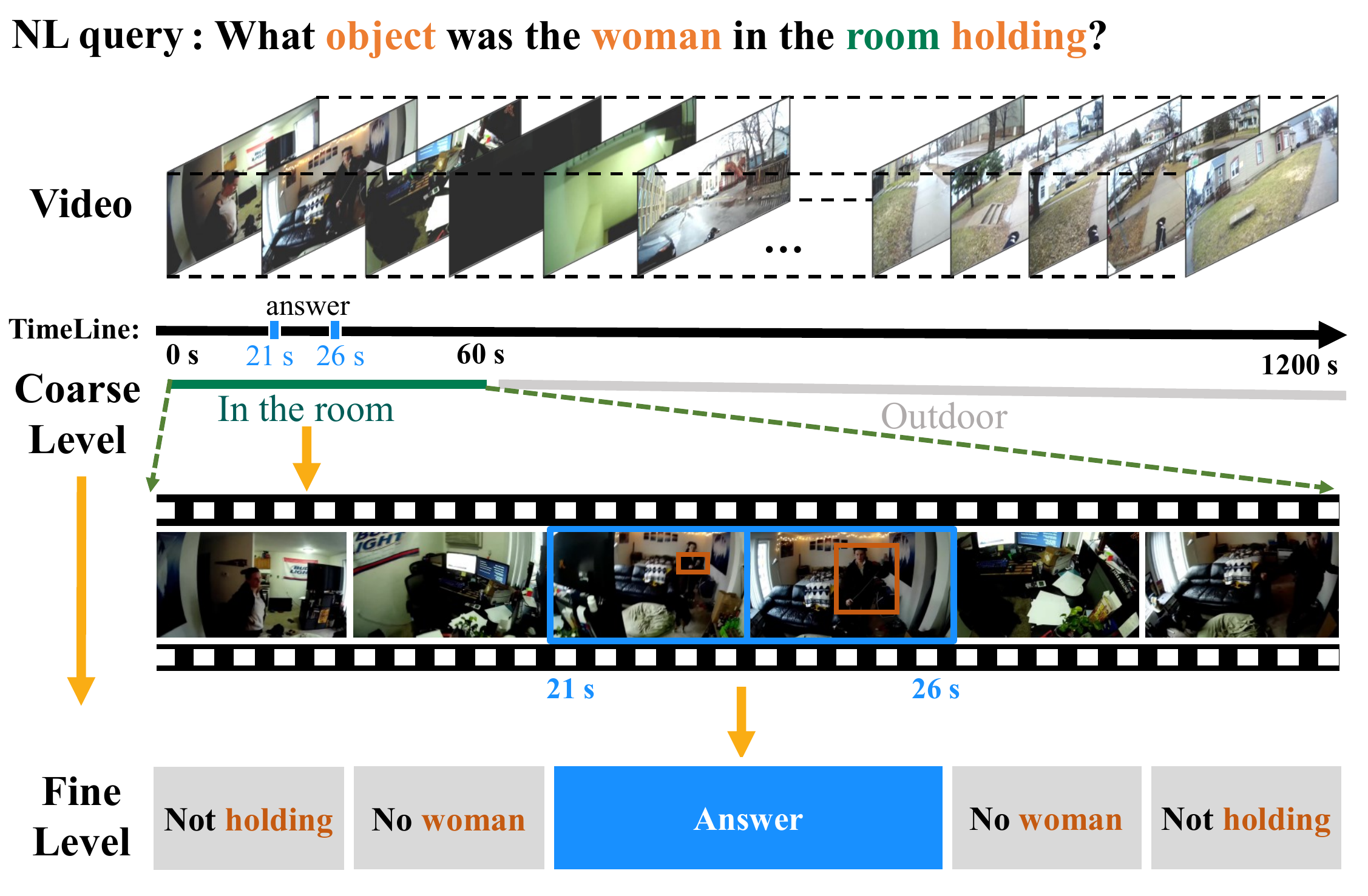}
\end{center}
\caption{An example of long video temporal grounding, which requires coarse-to-fine multi-modal alignment.}
\label{fig:intro_figure}
\vspace{-0.1in}
\end{figure}

Temporal grounding for \textbf{long videos} is especially highly-demanded and emerging due to the flourishing growth of online videos in quantity and length.
However, limited by previously available datasets~\cite{gao2017tall,krishna2017dense,lei2021detecting}, existing works~\cite{zhang2020span,zhang2020learning} mainly deal with relatively short videos ranging from 0.5 to 2.5 minutes on average. 
Recently, Ego4D~\cite{grauman2022ego4d} and MAD~\cite{soldan2022mad} datasets have been created and attempted to deal with long video, which spans from several minutes to hours.


Early attempts for long video grounding extends existing VTG models 
for short videos through sparse sampling~\cite{grauman2022ego4d} or a sliding window adaptation~\cite{soldan2022mad}. 
On the one hand, existing models for short videos typically downsample videos to a fixed-length frame sequence, which leads to \textit{temporal information loss} for long videos~(i.e., fewer visible frames via sparse sampling).
Besides, the window-based adaptation methods divide the long video into candidate windows and inference on every window, which leads to high inference computational cost.
On the other hand, massive moment candidates from long videos make their precise multi-modal alignment with the NL query more challenging,
which leads to \textit{contextual information loss}~(i.e., weaker alignment to fine-grained contents, like objects, actions, and scenes). As the motivating example shown in Fig. \ref{fig:intro_figure}, accurate grounding requires a coarse-grained localization of relevant video segments (e.g. ``\textit{in the room}" v.s. ``\textit{outdoor}"), and fine-grained alignment to detailed object and action in frames (e.g., ``\textit{women}" and ``\textit{holding}").
In conclusion, long video length poses two new challenges:
(1) higher computational cost during inference on the numerous windows of the entire long video;
(2) weaker multi-modal alignment due to the abundance of moment candidates.
To address these challenges, 
we propose \modelname, a COarse-to-fiNE alignment framework for long video temporal grounding. First, we slice the arbitrary long video into candidate windows. 
Then, we employ a query-guided window selection strategy for efficiency and further introduce a coarse-to-fine mechanism for effectiveness.
Specifically, the \emph{query-guided window selection} strategy efficiently reduces the sizeable window number of the long video to a modest amount via efficient alignment score computation.
Moreover, the \emph{coarse-to-fine} alignment mechanism consists of three modules to gradually align the multi-modal inputs via multi-scale granularity: 
(1) window (coarse granularity) selection to reliably select semantically relevant candidate windows; 
(2) window (coarse)-proposal (fine) joint contrastive learning to generate candidate proposals considering both inter-window and intra-window relations; 
(3) proposal (fine granularity) ranking to further accurately sort out the perfect proposal.
Notably, we incorporate contrastive learning into this mechanism. 
On the one hand, we utilize the powerful multi-modal alignment ability of contrastive vision-text pre-trained models~(e.g., CLIP), which computes matching scores between video frames and textual query for both window selection and proposal ranking stages.
On the other hand,  we select a contrasting negative window and design an inter-window contrastive loss during training for the joint learning stage.

With this coarse-to-fine design, \modelname~has the following advantages: (1) higher efficiency in handling long video inputs with less temporal information loss; (2) more accurate multi-modal alignment with less contextual information loss.
We evaluate \modelname~on two large-scale long video grounding benchmarks and achieve both the state-of-the-art results and significant performance boosts (3.13\%$\rightarrow$6.87\% on MAD, and 10.46\%$\rightarrow$13.46\% on Ego4D-NLQ in terms of R1@IoU=0.3).
Further analysis shows that the window selection
strategy shortens the inference speed by 2x for Ego4D-NLQ and 15x for MAD compared to inference on every window, without sacrificing its performance.
\paragraph{Contributions}
Our work presents two major contributions to the long video temporal grounding field. (1) We propose a novel coarse-to-fine alignment framework that utilizes a pipeline of \{window slicing and selection, proposal generation and ranking\}, resulting in state-of-the-art performance and high efficiency on two representative benchmarks. 
(2) We introduce a novel approach for incorporating contrastive learning into multi-modal alignment. 

%% file: Paragraphs/5-RelatedWork.tex
\section{Related Work}
\noindent\textbf{Video Temporal Grounding.}
Current models for this task typically fall into two categories based on the usage of proposals. \textit{Proposal-free methods} predict start/end timestamps directly, bypassing the generation of proposals~\cite{ghosh2019excl,Zeng_2020_CVPR,Zhang_2021_CVPR}.
Conversely, \textit{Proposal-based methods} merge proposal generation and ranking within an end-to-end framework~\cite{zhang2020learning,lei2021detecting,cao-etal-2021-pursuit}. An excellent survey paper provides further details on these models~\cite{zhang2022elements}.
However, these models are predominantly designed for relatively short videos, which leads to substantial information loss when adapted directly to long-form settings.
Regarding long-form video grounding, VSLNet-L~\cite{zhang2021natural} extends VSLNet~\cite{zhang2020span} with a multi-scale split-and-concat mechanism to address performance degradation. Still, the multi-scale mechanism adds computation cost during inference, and the sparse sampling approach suffers from significant temporal information loss for hour-long videos. 
Authors of Ego4D and MAD adapt 2D-TAN~\cite{zhang2020learning} and VLG-Net~\cite{soldan2021vlg} into simple window-based baselines, yet these models lack the ability for coarse-to-fine alignment, affecting the final results.

Furthermore, other studies have focused on different issues. For example, NaQ~\cite{ramakrishnan2023naq} addresses data scarcity with an effective data augmentation strategy, while DeNet~\cite{zhou2021embracing} tackles the ground-truth bias problem with a de-coupling and de-bias strategy.
There are also other related tasks involving language grounding with video corpus input~\cite{lei2020tvr,hou2021conquer} or spatial-temporal output~\cite{yang2022tubedetr}, but they fall outside the scope of this paper.

\begin{figure*}[!th]
\begin{center}
\includegraphics[width=\textwidth]{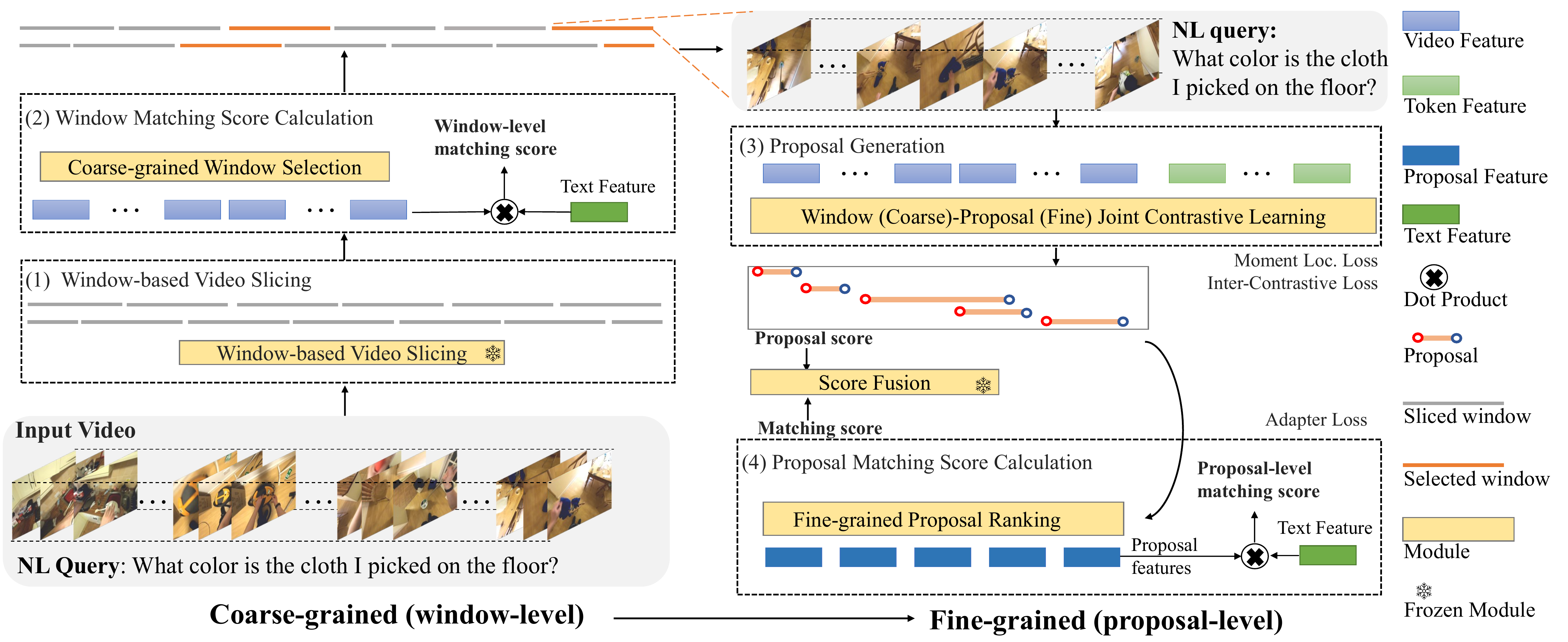}
\end{center}
\caption{Overview of \modelname. It  slices the long video into candidate windows~(\cref{sec:window}), selects semantic relevant windows~(\cref{sec:pre-filtering}), generates candidate proposals~(\cref{sec:contrast}), and ranks these proposals~(\cref{sec:fine-grained}).}
\label{fig:overview}
\vspace{-0.1in}
\end{figure*}

\noindent\textbf{Long-form Video Modeling.} Long-form video modeling is an emerging topic recently investigated in (multi-modal) video understanding, including classification, segmentation, localization, and retrieval.
The common challenges of long videos are modeling long-range temporal dependency,  efficiency issues, and accurate multi-modal alignment (if language is involved).  To ease long-range dependency issues, some works explore either feature memory banks~\cite{wu2019long, TallFormer_ECCV22}, tracked object-level representations~\cite{wu2021towards}, bayesian non-parametric model~\cite{qiu2023liveseg}, or structured state-space sequence layers~\cite{ViS4mer_ECCV22}. To alleviate efficiency issues, other works~\cite{ECLIPSE_ECCV22} exploit dense audio sampling as additional information to enable sparse video frame sampling.
In contrast, CONE mitigates both efficiency issues via query-guided window selection strategy and multi-modal alignment issues via novel contrastive learning incorporation.

\noindent\textbf{Contrastive Learning.} Contrastive learning has been widely studied in vision~\cite{misra2020self,he2020momentum}, language~\cite{gao-etal-2021-simcse,meng2021coco},
and multi-modal fields~\cite{liang2020learning,yan2021control,wang-etal-2022-contrastive-video}.  
Some recent VTG works also attempt to adopt contrastive learning. They mainly focus on frame-level contrastive loss via hard negative mining~\cite{zheng2022weakly} or  contrasting ground-truth frame with non-ground-truth frame~\cite{nan2021interventional,zhang2021video}. 
In contrast, we further propose window-level contrastive learning to repel the negative windows.
Moreover, CONE also jointly incorporates contrastive vision-text pre-trained models for accurate alignment, because those contrastive models show strong multi-modal alignment ability~\cite{luo2021clip4clip,xu-etal-2021-videoclip} derived from pre-training large-scale vision-text pairs.


%% file: Paragraphs/2-task.tex
\section{Task Definition}
\label{sec:task-definition}
Since never-ending video streams in real applications have a higher demand for long videos, in this paper, we study the task of video temporal grounding (VTG) in a more challenging setting with long video inputs. 
Taking a natural language (NL) query $Q$ and a long video $V$ as the inputs, the task of VTG  requires the system to locate the matched moment $M$ from the video $V$ relevant to the query $Q$. Formally, video $V=[v_1, v_2, \dots, v_{L_{v}}]$ is a sequence of uniformly sampled video frames, and $L_{v}$ denotes the length of the sampled frames. 
The NL query $Q=[q_1, q_2,\dots, q_{L_q}]$ is a sequence of tokens with sentence length $L_q$. 
The moment $M$ is a sub-sequence of $V$ that is relevant to $Q$.
The long-form video setting is more challenging because 
of the natural demands for more computation and time to process the entire video. Moreover, the accurate multi-modal alignment between each $v_i$ and $Q$ is also harder when $L_{v}$ increases.

%% file: Paragraphs/3-approach.tex
\section{Approach}
We present the proposed \modelname~for long video temporal grounding. 
As shown in Fig.~\ref{fig:overview}, we first slice the long video into several fixed-length video windows via a \textbf{sliding window} mechanism (\cref{sec:window}). 
Then, we propose a coarse-to-fine mechanism for efficient and effective multi-modal alignment.
At the coarse-grained window level, we introduce a \textbf{query-guided window selection} strategy (\cref{sec:pre-filtering}) to accelerate inference and select semantic relevant windows. At the window-proposal joint learning level, we rely on the existing VTG work for short videos to generate reliable candidate proposals and conduct both intra-window and proposed \textbf{inter-window contrastive learning} (\cref{sec:contrast}) to assign each proposal score.
At the fine-grained proposal level, we further accurately rank candidate proposals with a \textbf{fine-grained ranking} (\cref{sec:fine-grained}) mechanism.

\input{Paragraphs/3.1-window_slicing.tex}
\input{Paragraphs/3.2-pre-filtering.tex}
\input{tabs/data-statistic}
\input{Paragraphs/3.3-inter-contrastive.tex}

\input{Paragraphs/3.4-intra-fine-grained.tex}

%

%% file: Paragraphs/3.1-window_slicing.tex
\subsection{Window-based Video Slicing}
\label{sec:window}
To flexibly handle long videos without decreasing the sample rate and alleviate temporal information loss, we first slice the entire video $V$ into several video windows $W$. 
A sliding window with window length $L_w$ is used to be slid on the entire video to derive a set of $N_w$ fixed-length video windows $W_i=[v_{w^b_i+1},v_{w^b_i+2},...,v_{w^b_i+L_{w}}]$, where $w^b_i$ is the start index of window $i$.
Concretely, we slide the window by increasing $w^b$ with window stride $L_{w}/2$. 
Intuitively, not every window is correlated with the NL query, so we refer the \textit{positive window} to the window overlapping with the ground-truth moment, and the \textit{negative window} otherwise. 

%% file: Paragraphs/3.2-pre-filtering.tex
\subsection{Coarse-grained Window Selection}
\label{sec:pre-filtering}
Lengthy video input is sliced into a sequence of windows during inference. 
If the number of windows is $N_w$, the model needs to conduct the whole encoding-prediction process for $N_w$ times, which will become computationally costly with increased video length, especially when the model has enormous parameters.  
Therefore, it is necessary to reduce the inference computation cost by reliably filtering windows irrelevant to the natural language descriptions.
We propose a query-guided window selection strategy via contrastive vision-text pre-trained models. 
\paragraph{Vision-Text Contrastive Model.}
CLIP~\cite{radford2021learning} and EgoVLP~\cite{lin2022egocentric} are pre-trained with large-scale vision-text pairs via contrastive learning, aiming to align the visual representation with its related text representation. 
So they excel at multi-modal alignment, and efficient dot-product score computation provides higher matching efficiency. 
We utilize the pre-trained model to compute the video features $\bm{V}$ and the text features $\bm{Q}$ beforehand. 
 \begin{equation}
 \label{equ:features}
     \begin{aligned}
      \bm{V}&=[\bm{v}_1,\bm{v}_2,\dots,\bm{v}_{L_v}] \\
      \bm{Q}&=[\bm{q}_{\texttt{[CLS]}},\bm{q}_1,\bm{q}_2,\dots,\bm{q}_{L_q}]
     \end{aligned}
 \end{equation}
 where \texttt{[CLS]} is a special token at the beginning of text tokens. 
The multi-modal matching score $a_{j} = \bm{v}_j \cdot \bm{q}_{\texttt{[CLS]}}$ is computed via the efficient dot product between $j^{th}$ video feature and the text feature. And the window-level matching score $A_{i}$ is the maximum score of all the frames in window $i$: 
\begin{equation}
    A_{i} = \max([a_{w^b_i+1},a_{w^b_i+2},...,a_{w^b_i+L_w}])
\end{equation}

We rank all windows with $A_{i}$ and select the top-$k$ windows for inference. Thus, we reduce the number of candidate windows from $N_w$ to a constant $k$ to guarantee a controllable computation cost 
in an efficient and reliable manner.

%% file: tabs/data-statistic.tex
\begin{table*}[t]
\centering
\resizebox{1.0\textwidth}{!}{
    \begin{tabular}{lcccccccc}
    \toprule
    Dataset & Domain  & \#Query (train/val/test) & \#Video &   $N_{vocab}$    &     $\overline{L}_{video}$   &   $\overline{L}_{query}$    & $\overline{L}_{moment}$ & $\delta_{moment}$ \\
    \midrule
    Ego4D-NLQ & Open  & 11,291 / 3,874 /  4,005  & 998 / 328 / 333  &   3.3K    & 8.25min & 7.5   & 8.3s  & 2.8s \\
    MAD   & Movie & 280,183 / 32,064 / 72,044 & 488 / 50 / 112  & 61.4K & 110.77min & 12.7  & 3.9s & 3.2s \\
    \bottomrule
    \end{tabular}}%
     \caption{Statistics of two benchmarks. $N_{vocab}$  is word vocabulary size, $\overline{L}_{video}$
 is average video length,  $\overline{L}_{query}$  is average query word number, $\overline{L}_{moment}$ and $\delta_{moment}$ are the average and median ground-truth moment length.}
  \label{table:dataset_statistics}%
\vspace{-0.2in}
\end{table*}%

%% file: Paragraphs/3.3-inter-contrastive.tex
\subsection{Window (Coarse)-Proposal (Fine) \\ \quad \quad Joint Contrastive Learning}
\label{sec:contrast}
Since there are many excellent works in the VTG literature for short video input, we target \modelname~as a flexible plug-and-play framework on top of the existing proposed-based model. Existing models function as the base model to generate reliable candidate moment proposals for further processing.
In our scenario, the base model takes window $W$  and the NL query $Q$ as inputs and outputs several moment proposal candidates. Each of them has a moment proposal $(p_i)$ and corresponding score $(s_i)$, respectively.

Nevertheless, the training of the base model typically  focuses only on the positive window and conducts intra-window relation learning but neglects plenty of negative windows for long videos. In real practice,  negative windows are the majority during inference, which results in a discrepancy between training and inference.
To mitigate the discrepancy, we further design inter-window contrastive learning to distinguish the positive and negative windows.
As a result, the overall training loss consists of two parts: (1) the intra-window loss of the base model, and (2) the proposed inter-window contrastive loss.

Concretely, each training instance has both a positive window and a random contrasting negative window from the long video. 
We discriminate the negative window from the positive window through proposal-level comparison. Proposals in the negative window should be assigned with minimized scores compared with positive proposals~(e.g., the IoU with ground truth is large than 0.7) in the positive window, as follows,
\begin{equation}
\mathcal{L}_{con} =  - \sum_{p_i^{+} \in W^{+}} \log (s_i) - \sum_{p_i^{-} \in W^{-}} \log (1 - s_i)
\end{equation}
where ${L}_{con}$ is the proposed contrastive loss, $p_i^{+}$ is each positive proposal from the positive window $W^{+}$ and $p_i^{-}$ is each proposal from the negative window $W^{-}$, $s_i$ is the corresponding proposal score.


This mechanism can be generalized to different existing proposal-based models with different intra-window losses. To show the generalization ability of~\modelname, we test on two kinds of base models: 2D-TAN \cite{zhang2020learning} and Moment-DETR\cite{lei2021detecting} due to their available codes and superior performances. Please refer to Appendix~\ref{sec: training-loss} for comprehensive training details.

%% file: Paragraphs/3.4-intra-fine-grained.tex
\subsection{Fine-grained Proposal Ranking}
\label{sec:fine-grained}
With the increased length of video inputs, the fine-grained attention between each video frame and the text query will be weakened by many other perturbed frames, resulting in \textit{contextual information loss}.
To remedy this issue, we propose a fine-grained ranking strategy to conduct accurate  proposal ranking utilizing  proposal matching scores computed by contrastive vision-text pre-trained models (described in \cref{sec:pre-filtering}).
Note that we simply re-use the pre-computed video frames and text query features as in Eq. \eqref{equ:features}. 
\paragraph{Visual Adapter.}
With a lightweight visual adapter on top of CLIP, we exploit adapter-based tuning to adapt the representations from the general contrastive model to the data distribution of the current downstream task.
Inspired by \citet{gao2021clip}, our main idea is to add an additional bottleneck layer to learn the task-adaptive visual features and conduct residual-style blending with the original pre-trained
features. 
The lightweight adapter complies with a 2-layer FFN followed by ReLU. The $i^{th}$ adapted visual feature is: 
$\hat{\bm{v}_i} = \textrm{Adapter}(\bm{v}_i) + \bm{v}_i$.
Then, the \textbf{proposal feature} for the $j^{th}$ proposal is computed with the mean pooling of all the adapted video features in it: $\bm{h}_j=\textrm{Mean}([\hat{\bm{v}_{b_j}},\dots,\hat{\bm{v}_{e_j}}])$

For adapter training, we denote the positive proposal (with feature $\bm{h}_{pos}$) as the ground-truth one, and the negative proposals are the other in the same batch. We follow the standard contrastive learning and use the NCE loss:
\begin{equation}
    \mathcal{L}_{adapt} = - \sum_{pos} ( log \frac{\exp(\bm{h}_{pos}\cdot \bm{q}_{\texttt{[CLS]}})}{\sum_{j}\exp(\bm{h}_{j}\cdot \bm{q}_{\texttt{[CLS]}})} )
\label{equ: adapt}
\end{equation}

Note that we also use the adapted visual feature to compute the window-level matching score in \cref{sec:pre-filtering} to improve window selection quality.

\paragraph{Ranking Score Computation.}
Finally, we aim to conduct a fine-grained ranking for proposals. For the $j^{th}$ proposal, the final ranking score is fused with two components: (1) proposal scores $s_j$ generated from the previous module and (2) fine-grained matching scores $m_j$  computed by adapted CLIP-based proposal feature: $m_j=\bm{h}_j\cdot\bm{q}_{\texttt{[CLS]}}$. 
The former captures the correlation between proposals via the sophisticated contextual model design, while the latter focuses on fine-grained content matching between frames in the proposal and the textual query. 
We perform min-max normalization for these two types of scores for a more stable score fusion. 
The final ranking score $r_{j}$ is the sum of two normalized scores:
\begin{equation}
\label{eq:neg_loss}
\begin{aligned}
 \tilde{s_{j}} = &~\textit{MinMax}([s_1,s_2,...,s_{N_p}]) ,\\
 \tilde{m_{j}} = &~\textit{MinMax}([m_1,m_2,...,m_{N_p}]) \\ 
 r_{j} = &~\tilde{s_{j}} + \tilde{m_{j}} \\
\end{aligned} 
\end{equation}
where $N_p$ is the total candidate proposal number.

%% file: Paragraphs/4-experiment.tex
\input{tabs/main-result.tex}
\section{Experiments}

We conduct experiments to explore the effectiveness of \modelname~from the following aspects: (1) model comparison with SOTA methods (\cref{sec:main}); (2) ablation study to analyze the impact of each component and different variants (\cref{sec:modal-analysis}); (3) efficiency analysis of acceleration with window selection (\cref{sec:efficiency}) and (4) qualitative analysis (\cref{sec:case-study}). Implementation details are given in \textbf{Appendix~\ref{sec: implementation}}.

\input{Paragraphs/4.1-dataset.tex}
\input{Paragraphs/4.2-setting.tex}

\input{Paragraphs/4.3-baseline.tex}

\subsection{Model Analysis}
\label{sec:modal-analysis}
\input{tabs/ablation}
Ablation studies are shown in table \ref{tab:ablation} to unveil the effectiveness of each component in \modelname. 
We consider three components: 
(a) \textbf{contrastive loss} (\cref{sec:contrast}), which is eliminated by only training on positive windows without our contrastive loss; 
(b) \textbf{fine-grained ranking fusion} (\cref{sec:fine-grained}) can be removed by taking only the proposal score from the proposal generation module for ranking; 
(c) \textbf{visual adapter} (\cref{sec:fine-grained}) is removed by using general CLIP-based features for matching score computation. 
Note that the full \modelname~model refers to the first row, and the baseline model (i.e., the window-based adaptation of Moment-DETR) refers to the last row.

From the table~\ref{tab:ablation}, we highlight the following findings for Ego4D-NLQ dataset~(MAD shows similar trends): 
(1) Ablating visual adapter leads to a performance drop from 14.15\% to 12.62\%, which indicates that domain-adaptation of visual features is essential in modelling task-specific semantic variance. (2) Eliminating fine-grained ranking also harms the performance (row 2 vs. row 3), showing that capturing fine-grained semantic alignment benefits accurate grounding.  (3) Further removing contrastive loss (row 3 vs. row 4) leads to a significant performance drop of 3.36\%, which reveals that identifying the inter-window semantic variance is critical for reliable grounding.

\begin{figure}[!h]
\begin{center}
\includegraphics[width=0.99\linewidth]{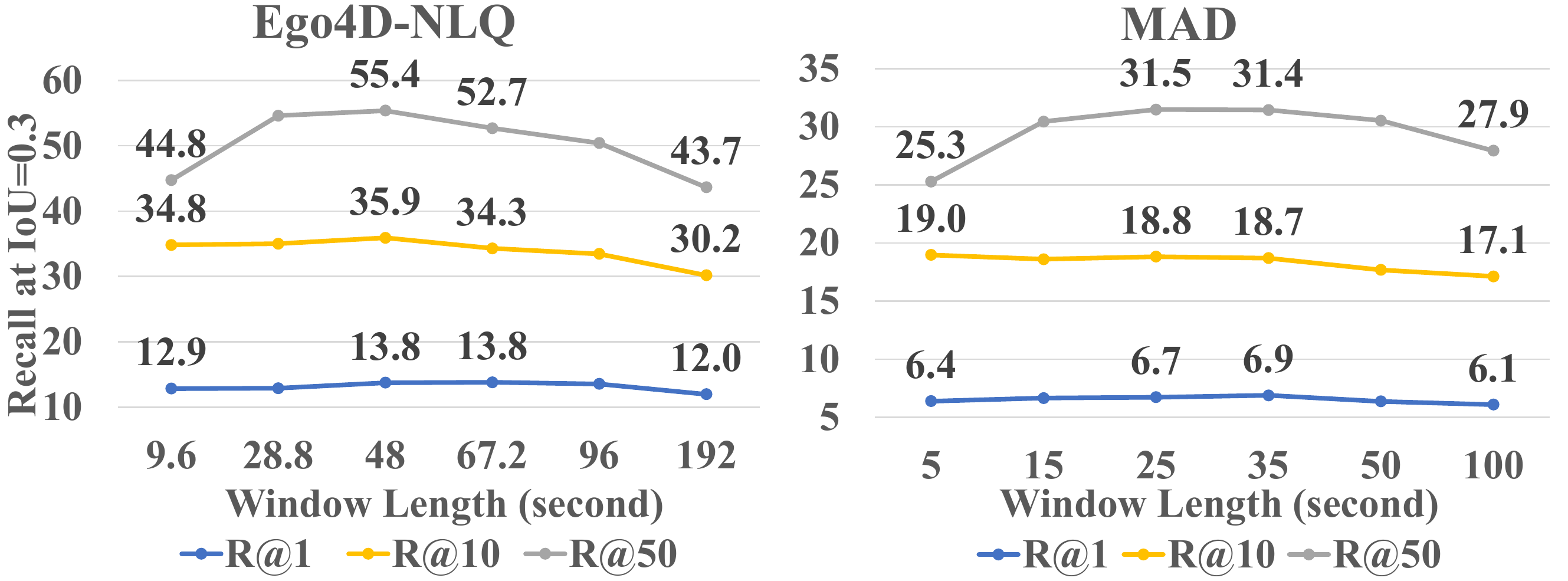}
\end{center}
\caption{Performance (Recall k@IoU=0.3) on both datasets w.r.t different window lengths. }
\label{fig:window_length_vs_performance}
\vspace{-0.2in}
\end{figure}

\paragraph{Influence of window length.}
Figure~\ref{fig:window_length_vs_performance} exhibits how different window lengths affect the overall performance of \modelname. 
We observe that increasing the window length indeed brings performance variance. Longer window length (more visible frames in a single window) can model the interaction between more frames, but can also weaken the multi-modal attention between the NL query and every single frame as the performance drops significantly with the largest window size. We find a better trade-off between window length and performance, i.e., nearly 48 seconds ( 90 video features) for Ego4D-NLQ and nearly 25 seconds (125 video features) for MAD through this analysis.

\input{Paragraphs/4.5-efficiency.tex} 

\input{Paragraphs/4.6-qualitative.tex}



%% file: tabs/main-result.tex
\begin{table*}[!t]
\centering
 \resizebox{1.0\textwidth}{!}{
    \begin{tabular}{lccccrccccrcccc}
    \toprule
    \multirow{2}[4]{*}{Model} & \multicolumn{4}{c}{IoU = 0.1} &       & \multicolumn{4}{c}{IoU = 0.3} &       & \multicolumn{4}{c}{IoU = 0.5} \\
\cmidrule{2-5}\cmidrule{7-10}\cmidrule{12-15}          & R1    & R5    & R10   & R50   &       & R1    & R5    & R10   & R50   &       & R1    & R5    & R10   & R50 \\
    \midrule
    2D-TAN$\ast$  & 3.22   &  11.90   &  19.01   &  39.63   &      & 2.52  & 9.25 & 14.72  & 32.65 &    & 1.58  & 5.69 & 9.06 &  22.46   \\
    Moment-DETR$\ast$  &  3.60  & 12.98   &  20.70   & 41.32  &  & 2.81     &  9.86   &  15.53 & 32.02  &    &  1.67  & 5.58  & 8.68  &  17.97   \\
     VLG-Net & 3.64  & 11.66 & 17.39 & 39.78 &       & 2.76  & 9.31  & 14.65 & 34.27 &       & 1.65  & 5.99  & 9.77  & 24.93 \\
    CLIP  & 6.57  & 15.05 & 20.26 & 37.92 &       & 3.13  & 9.85  & 14.13 & 28.71 &       & 1.39  & 5.44  & 8.38  & 18.80 \\
     \midrule
    \modelname~(2D-TAN)  & 8.30    &  \textbf{20.52}  &  26.27    &  36.61    &      &  6.43  & \textbf{16.46}   &  \textbf{21.83} &  33.21  &   & 3.55   &   \textbf{10.44} & \textbf{15.07}   &  \textbf{26.49}    \\
    \modelname~(Moment-DETR)    & \textbf{8.90}  & 20.51 & \textbf{27.20} & \textbf{43.36} &       & \textbf{6.87}  & 16.11 & 21.53 & \textbf{34.73} &       & \textbf{4.10}  & 9.59  & 12.82 & 20.56 \\
    \bottomrule
    \end{tabular}}%
    \caption{Performance on the test set of MAD dataset. We implement the models with $\ast$ and report their performances.}
   \label{tab:mad-test}%
   \vspace{-0.1in}
\end{table*}%


\begin{table}[!ht]
\centering
    \resizebox{0.48\textwidth}{!}{
    \begin{tabular}{l| l|cccc}
    \toprule
    \multirow{2}[2]{*}{Model} &   \multirow{2}[2]{*}{Feature} & \multicolumn{2}{c}{\textbf{IoU = 0.3}} & \multicolumn{2}{c}{\textbf{IoU = 0.5}} \\
         &     & R1    & R5    & R1    & R5 \\
    \midrule
   \multicolumn{6}{c}{Validation split} \\
    \midrule
    2D-TAN &  SF + B & 5.04 & 12.89 & 2.02 & 5.88  \\ 
    Moment-DETR$\ast$ & SF + B & 5.09  & 17.17 & 2.99 & 9.16  \\
    2D-TAN$\ast$ & EgoVLP & 7.10 & 18.28  & 3.82 & 11.00  \\ 
  Moment-DETR$\ast$ & EgoVLP & 8.23 & 23.23 & 5.01 & 13.37 \\
    VSLNet  & EgoVLP & 10.84 & 18.84 & 6.81 & 13.45  \\ 
    \midrule
    CONE (2D-TAN) & EgoVLP & 11.00  & 25.06 & 6.09
     & 15.51  \\ 
    CONE (Moment-DETR) &  SF + B & 10.64 & 24.47 & 5.76  & 13.32  \\ 
    CONE (Moment-DETR) & EgoVLP & \textbf{14.15} & \textbf{30.33} & \textbf{8.18}  & \textbf{18.02} \\
    \midrule
    \multicolumn{6}{c}{Test split} \\
    \midrule
    2D-TAN$\ast$ & EgoVLP & 6.89 & 14.86 & 3.80 & 8.47  \\ 
    Moment-DETR$\ast$ & EgoVLP & 9.14 & 18.66 & 5.05 & 10.59  \\ 
    VSLNet  & EgoVLP & 10.46 & 16.76 & 6.24 & 11.29 \\
    \midrule
    CONE (2D-TAN) & EgoVLP & 11.08 & 19.96 & 5.87 &   11.44\\ 
    CONE (Moment-DETR) & EgoVLP &  \textbf{13.46}     &  \textbf{23.68}     & \textbf{7.84}      & \textbf{14.16}  \\
    \bottomrule
     \end{tabular}}%
  \caption{Performance on the Ego4D-NLQ dataset. SF + B is short for the SlowFast and Bert features. We implement the models with $\ast$ and report their performances. }
  \label{tab:ego4d}%
\end{table}

%% file: Paragraphs/4.1-dataset.tex
\subsection{Dataset}
\label{sec:dataset}

We conduct comprehensive experiments on two representative large-scale benchmarks on long video temporal grounding:  Ego4D-NLQ~\cite{grauman2022ego4d} and MAD~\cite{soldan2022mad}.
The data statistics are summarized in Table~\ref{table:dataset_statistics}. 

\textbf{Ego4D-NLQ} is a subtask of the Ego4D dataset.  
Ego4D is a large-scale egocentric video understanding benchmark, where 931 camera wearers worldwide record their daily activities in hundreds of scenarios.
The unedited videos involved have variant lengths ranging from 3.5 min. to 20 min. 
The NL query is designed to retrieve the relevant moment from the episodic memory of camera wearers, and involves 13 question types for locating different types of information.

\textbf{MAD} is a large-scale long video temporal grounding benchmark on full-length movie videos. The video duration is magnificent longer than previous benchmarks at an average length of 110.8 min~(ranging from 47 min. to 202 min).
The textual queries in the training set are derived from translated movie audio descriptions from professional narrators. The annotations in the evaluation set are of higher quality derived from the LSMDC dataset~\cite{rohrbach2017movie} with more strict manual refinement. The NL queries are cleaner and the ground-truth moments have preciser temporal boundaries. 

%% file: Paragraphs/4.2-setting.tex
\subsection{Experimental Settings}
\label{sec:exp-setting}
\subsubsection{Evaluation Metric.} 
For consistent comparison,
we follow the previous baseline evaluation setting and use the metric $\textrm{Recall}@\mathit{k}$ at $\textrm{IoU}=\theta$~(R@$\mathit{k}$). 
It is the percentage of queries, having at least one prediction among the top-$k$ predictions, whose temporal IoU with ground-truth is larger than the threshold $\theta$ (0.3 or 0.5). 
Note that there is only one ground-truth answer for each query in both datasets. 

\subsubsection{Visual and Textual Features.}
For MAD dataset, we use CLIP \cite{radford2021learning} to extract the visual and textual features, which are then utilized in the inputs of the window selection, proposal generation, and fine-grained ranking stages. The MAD authors extract the visual features every 0.2s~(5fps). 
For Ego4D-NLQ dataset, 
we adopt the egocentric contrastive pre-trained model, i.e., EgoVLP~\cite{lin2022egocentric}, to extract visual and textual features, because CLIP is trained with third-person image-text pairs and has the domain adaptation gap for first-person videos. We extract the visual feature every 16 frames~(1.875fps).  

%% file: Paragraphs/4.3-baseline.tex
\subsection{Model Comparison}
\label{sec:main}
\subsubsection{Baselines.}
We compare \modelname~to the following methods: 
(1) the window-based adaptation of proposal-based models, i.e., \textbf{2D-TAN}~\cite{zhang2020learning}, \textbf{VLG-Net}~\cite{soldan2021vlg}  and   \textbf{Moment-DETR}~\cite{lei2021detecting};
(2) the sparse sampling-based proposal-free model 
\textbf{VSLNet}~\cite{zhang2020span};
(3) the two-stage model \textbf{CLIP}~\cite{radford2021learning}, which first generates offline proposals and then ranks those proposals with CLIP matching scores.

\paragraph{Results on Ego4D-NLQ.}
Table \ref{tab:ego4d} reports the performance comparison on the validation and the blind test set of Ego4D-NLQ. Regarding all the metrics, \modelname~outperforms these baselines by a large margin. 
Take R1@IoU=0.3 and R5@IoU=0.3 as examples, the absolute performance gains are +3.31\% and +11.49\% on the val. set, and +3\% and +6.92\% on the blind test set, respectively. 
CONE also achieves consistent performance gains using different features and base models, showing a better generalization ability.
\paragraph{Results on MAD.}
The main results of MAD 
are shown in Table \ref{tab:mad-test}, and demonstrate that \modelname~ achieves state-of-the-art performances with obvious boost (e.g., 3.13\%$\xrightarrow{\text{+119\%}}$6.87\% and 9.85\%$\xrightarrow{\text{+63.5\%}}$16.11\% improvement on R1@IoU=0.3 and R5@IoU=0.3).

From the two tables, 
the SOTA performance on two benchmarks demonstrates the effectiveness of \modelname.
We speculate the outstanding results are brought by the following reasons: (1) \modelname~processes the entire video without decreasing sample rate, which alleviates temporal information loss; (2) the coarse-to-fine mechanism enables the better alignment ability of relevant proposals to the NL query and reduces contextual information loss.
The lower results on the MAD benchmark also verify that the grounding task indeed becomes more challenging with longer videos.

Regarding base model selection, we observe that, compared with Moment-DETR,  the performance of 2D-TAN is worse in Ego4D-NLQ but comparable in the MAD dataset. This gives the clues that the base model might not be influential to the final results given more training samples for hour-level videos.
Because of the lower parameter and GLOPs number of Moment-DETR, from now on, \modelname~adopts it as the default base model.


%% file: tabs/ablation.tex
\begin{table}[htbp]
\centering
    \resizebox{0.48\textwidth}{!}{\begin{tabular}{l|c|c}
    \toprule
    \multicolumn{1}{p{8.93em}|}{Model} & \multicolumn{1}{l|}{Ego4D-NLQ} &
    \multicolumn{1}{l}{MAD} \\
    \midrule
    \multicolumn{1}{p{8.93em}|}{CONE} & 14.15 & 6.91 \\
    \hspace{3mm} w/o visual adapter & 12.62 & 6.73 \\
    \hspace{6mm} w/o fine-grained ranking & 11.59 & 4.66 \\
    \hspace{9mm} w/o contrastive learning & 8.23  & 2.56 \\
    \bottomrule
    \end{tabular}
    }
    \caption{Cumulative ablation study on the val. set of Ego4D-NLQ and MAD datasets, taking \textbf{R1@IoU=0.3} as the metric.}
  \label{tab:ablation}%
  \vspace{-0.15in}
\end{table}%

%% file: Paragraphs/4.5-efficiency.tex
\subsection{Efficiency Analysis}
\begin{figure}[!ht]
\begin{center}
\includegraphics[width=0.99\linewidth]{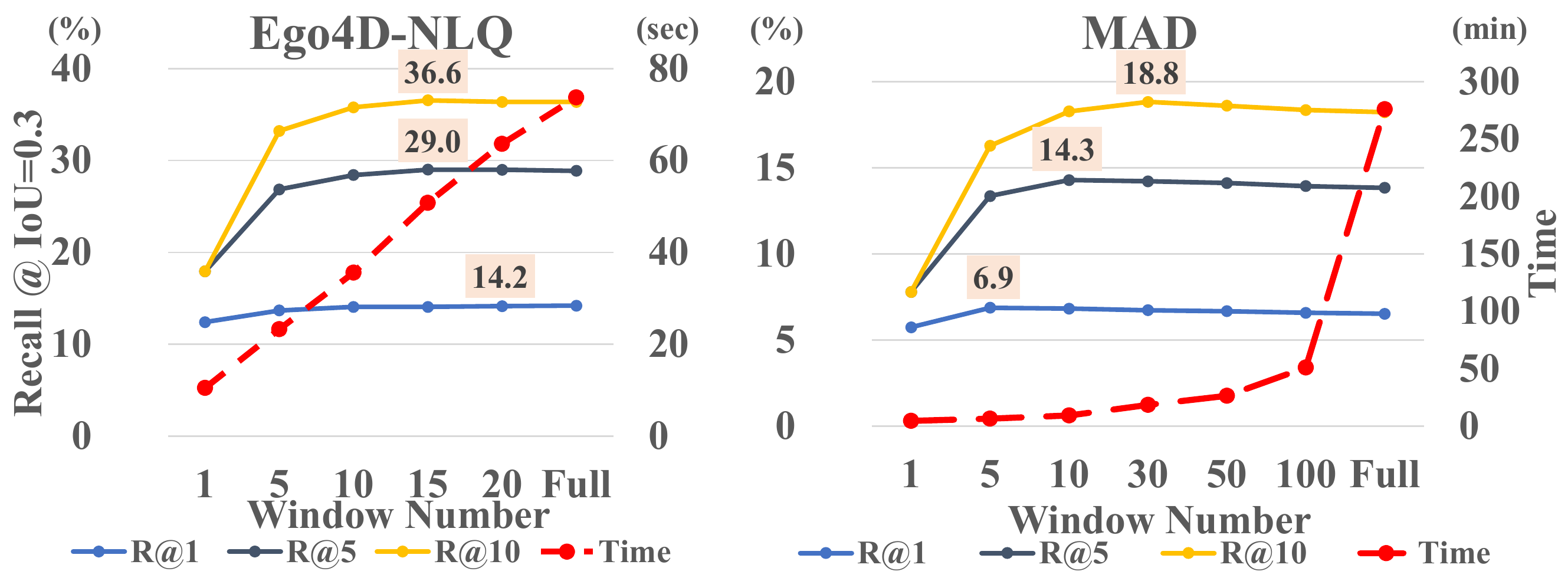}
\end{center}
\caption{Inference time w.r.t window numbers is denoted by \textbf{red dashed} lines for efficiency analysis. 
Performance (Recall k@IoU=0.3) w.r.t window numbers is shown with solid lines.}
\label{fig:window_number_vs_speed}
\vspace{-0.1in}
\end{figure}

\begin{figure*}[!ht]
\begin{center}
\includegraphics[width=0.99\linewidth]{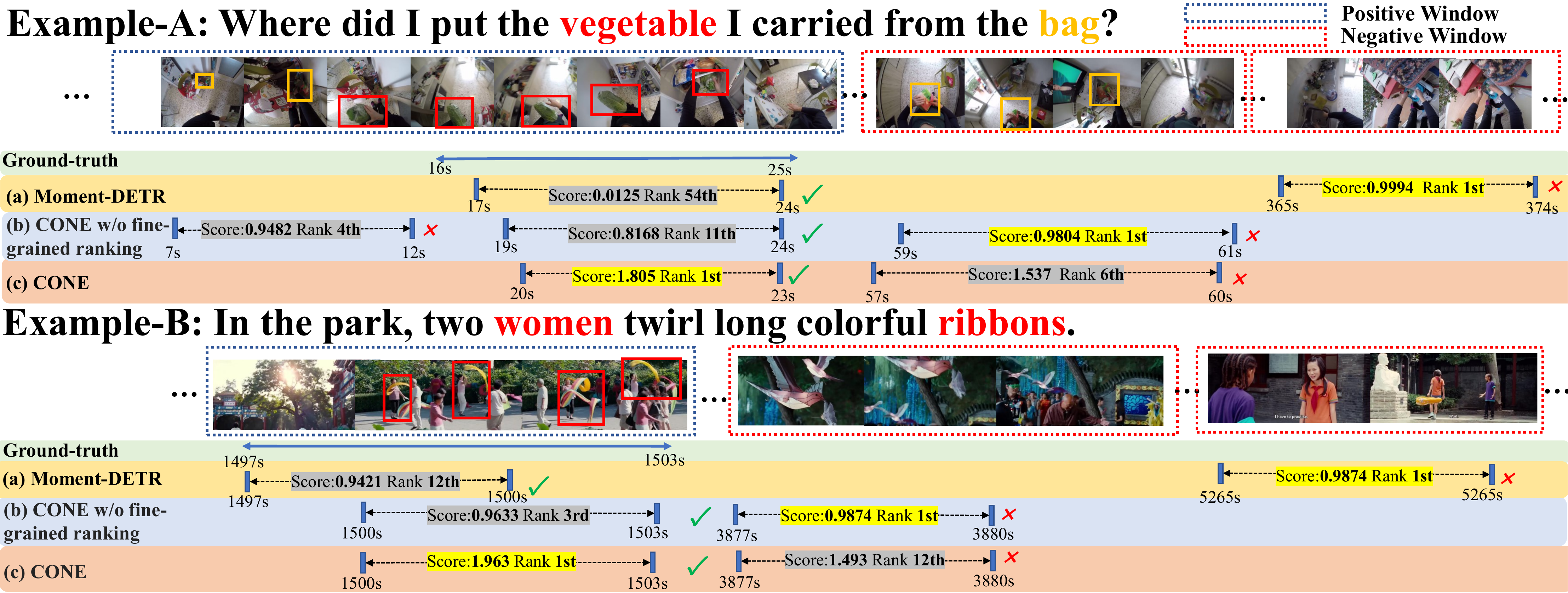}
\end{center}
\caption{ Two successful examples. Example-A and Example-B come from the Ego4D-NLQ and MAD datasets, respectively. We compare three settings: (a) baseline Moment-DETR; (b) CONE w/o fine-grained ranking and (c) the full CONE. The major difference between (a) and (b) is \textbf{inter-window contrastive learning}.}
\label{fig:qualitative}
\vspace{-0.15in}
\end{figure*}

\label{sec:efficiency}
Figure~\ref{fig:window_number_vs_speed} shows inference speed with respect to different window numbers after the window selection stage on the val. sets of Ego4D-NLQ and MAD. 
We observe that the overall inference time is approximately linearly reduced with smaller window numbers, and the performance of \modelname~is relatively stable when the filtered window number becomes 10 for both datasets.
Surprisingly, we also observe that \modelname~achieves the optimal performance~(shown as the box with value in Fig.~\ref{fig:window_number_vs_speed}) at a modest size of window number rather than the full size. Those observations show that our query-guided window selection strategy can reliably filter the irrelevant windows and largely improve the efficiency of long video temporal grounding.

Concretely, the average window numbers for full videos (before filtering) are 23.3 and 588 for Ego4D-NLQ and MAD, respectively.
If we set the filtered window number to 10 for Ego4D-NLQ and 30 for MAD (better trade-off values obtained from Fig. \ref{fig:window_number_vs_speed}), the total numbers of windows for inference are reduced by 2.3x and 19.6x, with a marginal performance variance (R@1) by -0.1\% and +0.2\% for Ego4D-NLQ and MAD benchmarks, respectively.
In real practice, we consider the effect of implementation methods and window selection time. When \modelname~is inferred using one P100 GPU, its running time (w/o feature extraction and post-processing time, because both time costs are the same for all methods) is largely reduced (74 s$\xrightarrow{\text{-2x}}$36 s and 276 min.$\xrightarrow{\text{-15x}}$18 min.) on the val. set of Ego4D-NLQ and MAD.

Furthermore, we also conduct an inference comparison between VSLNet and CONE on Ego4D-NLQ. Using the official Ego4D released code, VSLNet takes approximately 12 seconds to infer the Ego4D validation split and achieves a Recall@1 score of 10.84 at IoU=0.3. In contrast, CONE, with the extreme setting of selecting only the top 1 window, takes about 10.5 seconds to infer the Ego4D validation split and achieves a Recall@1 score of 12.4 at IoU=0.3. It demonstrates that CONE strikes an optimal tradeoff between accuracy and speed compared to sparse sampling-based baselines. 

%% file: Paragraphs/4.6-qualitative.tex
\subsection{Qualitative Analysis}
\label{sec:case-study}
Fig. \ref{fig:qualitative} shows two success examples to analyze the effect of \textbf{contrastive learning} and \textbf{fine-grained ranking}. We observe that baseline Moment-DETR (a) is not capable of repelling negative windows  and tends to give an equally high score to the proposal in the negative window, thus it wrongly ranks the correct prediction to a much lower position~(e.g., $54^{th}$ in Example A). Further adding inter-window contrastive learning mechanism (b), the rank position of the ground truth moment is somewhat improved, but it still lacks fine-grained matching to detailed contents to push the perfect proposal into the first place. For example, in Query-A, the most essential object is the vegetable rather than the bag;  In Query-B, the query requires the precise scene ``in the park" and the objects ``two women" and  ``ribbons". Finally, adding fine-grained ranking (c), the full \modelname~successfully locates the ground-truth moment relevant to the textual query. More qualitative examples with both success and failure cases are shown in Appendix~\ref{sec: more-example}.

%% file: Paragraphs/6-conclusion.tex
\section{Conclusions}
We present \modelname~, a COarse-to-fiNE alignment framework for long video temporal grounding. 
\modelname~jointly achieves state-of-the-art performances on two benchmarks and high efficiency while keeping high frame sample rate.
Through our experiments,
we show the proposed coarse-to-fine mechanism via contrastive learning plays an important role in boosting performance.
Regarding efficiency, the introduced query-guided window selection strategy largely accelerates inference by 2x and 15x on Ego4D-NLQ and MAD benchmarks, respectively, with even slight performance gains. 
Since \modelname~ can be generalized with different existing proposal-based models, we hope it can be used to improve the model efficiency and performance for temporal grounding on long videos. 

%% file: Paragraphs/limitation.tex
\section*{Limitations}
Currently, \modelname~ is mainly implemented for proposal-based models as they can generate explicit moment proposals for the introduced inter-window contrastive learning. 
In contrast, proposal-free methods (e.g., VSLNet~\cite{zhang2020span}) directly predict the start/end timestamps without explicit proposals. 
In the future, it is worthwhile to explore how to incorporate the coarse-to-fine alignment mechanism with proposal-free methods to enhance the generalization ability of \modelname.

Furthermore, \modelname~falls short on the ground-truth moment case whose duration is longer than the adopted video window duration. 
To ease this issue, future work can explore adaptive-duration window slicing to ensure complete containment of scenes and events within windows or some rule-based proposal merging techniques.

%% file: Paragraphs/appendix.tex
\newpage
\begin{figure*}[!ht]
\begin{center}
\includegraphics[width=0.99\linewidth]{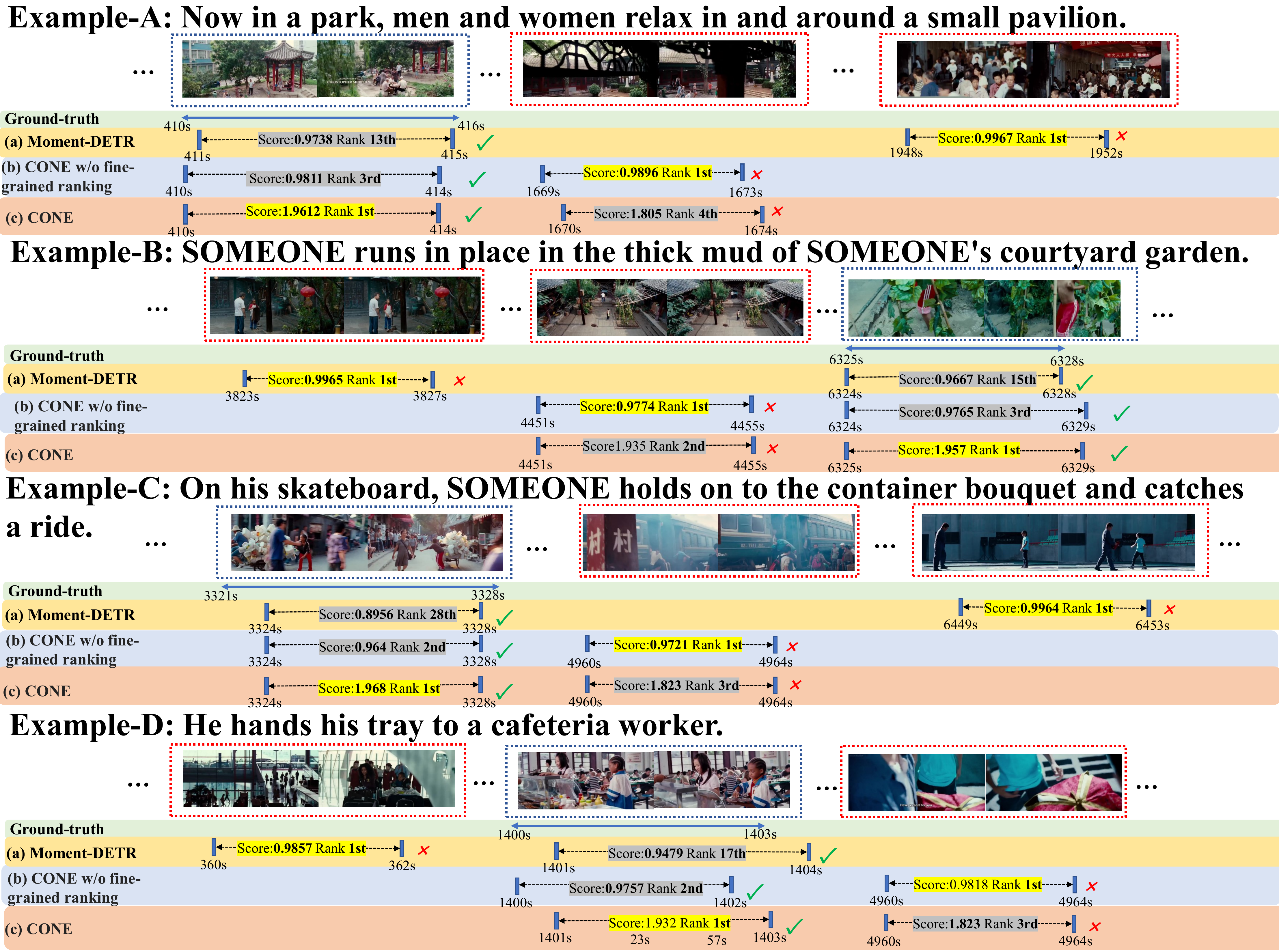}
\end{center}
\caption{ Four success examples on the MAD dataset. }
\label{fig:appendix_success}
\end{figure*}
\section{Model Details}
\subsection{Implementation Details}
\label{sec: implementation}
During training, we perform parameter optimization via AdamX and  set the learning rate to 1e-4 for the base model and 1e-5 for the visual adapter. We set the batch size to 32 and adopt the early stopping strategy. During inference, we use Non-Maximum Suppression~(NMS) with a threshold of 0.5 as post-processing.

\begin{figure*}[!t]
\begin{center}
\includegraphics[width=0.99\linewidth]{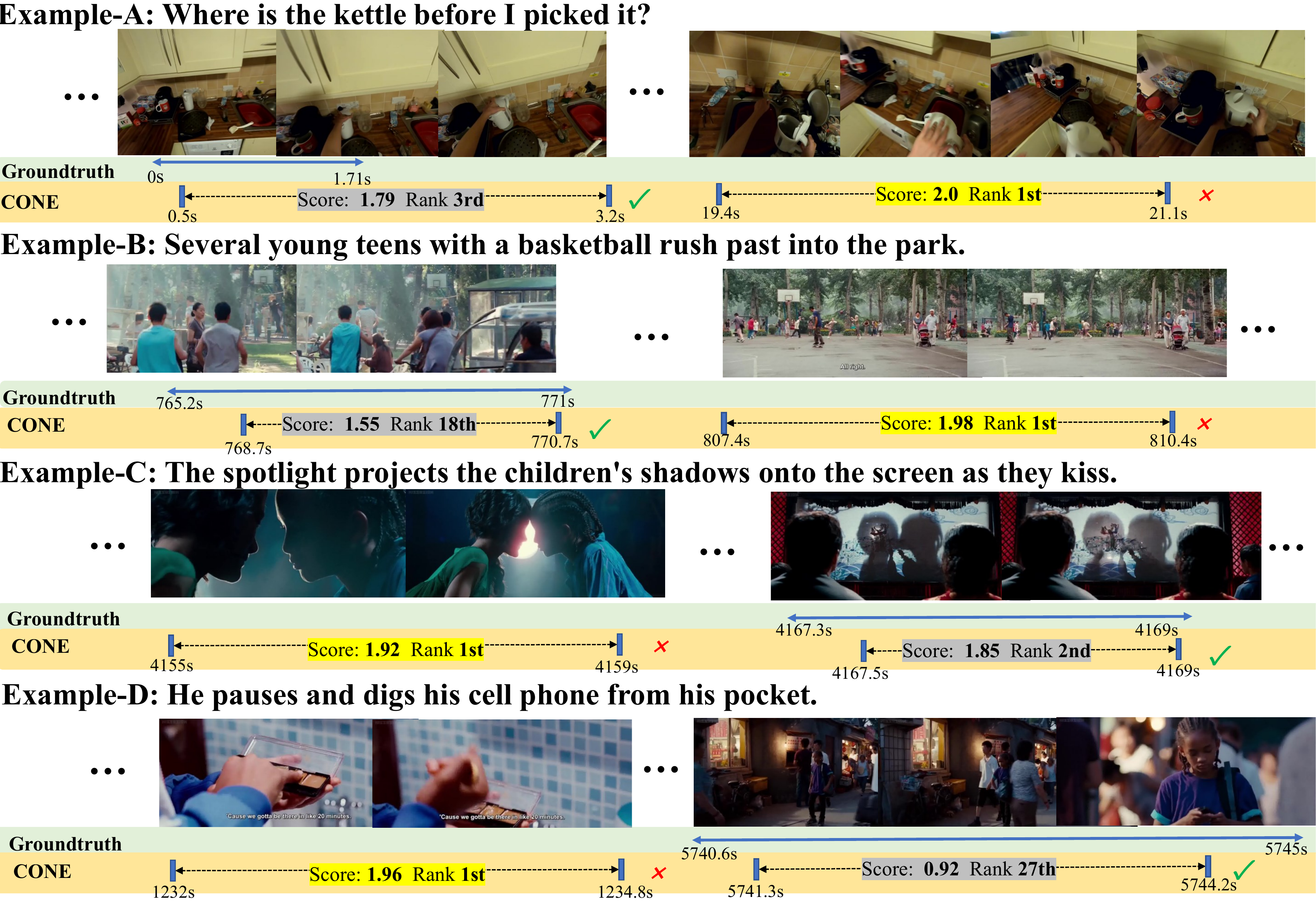}
\end{center}
\caption{ Four failure examples. Example-A comes from the Ego4D-NLQ dataset; the remaining three examples are from the MAD dataset.}
\label{fig:appendix_failure}
\end{figure*}

\paragraph{Moment-DETR Base Model.} The experiments are conducted on one P100 GPU.
For the hyperparameters of network architecture, we set the hidden size $d$ to 256, the transformer layer number in the encoder/decoder to 2, and the moment query number to 5. As a result, the total parameters of CONE are 4.35M (4.22M Moment-DETR + 0.13M visual adapter).
We set the window length to 90 video features~(48 seconds)  and 125 video features~(25 seconds) for the Ego4D-NLQ and MAD datasets, respectively.  
We train Ego4D-NLQ for 150 epochs and MAD for 30 epochs, and the training time is about 3 hours for the Ego4D-NLQ and 18 hours for the MAD dataset. 
During inference, we set the filtered window number to 10 and 30 for the Ego4D-NLQ and MAD datasets, respectively. 

\paragraph{2D-TAN Base Model.} The experiments are conducted on two V100 GPUs.
For the hyperparameters of network architecture, we set the hidden size $d$ to 256, the convolution network layer to 4, and the kernel size to 9.
As a result, the total parameters of CONE are 23.99M (23.86M 2D-TAN + 0.13M visual adapter).
We set the window length to 64 video features~(34.1 seconds) and 128 video features~(25.6 seconds) for the Ego4D-NLQ and MAD datasets, respectively.   We train Ego4D-NLQ for 90 epochs and MAD for 6 epochs, and because of 6x more model parameters than Moment-DETR, the training time is quite larger, i.e., about 18 hours for the Ego4D-NLQ and 3 days for the MAD dataset. 
During inference, we set the filtered window number to 5 and 15 for the Ego4D-NLQ and MAD datasets, respectively.

\subsection{Training Loss Details}
\label{sec: training-loss}

\paragraph{Moment-DETR Base Model.} The original loss of Moment-DETR consists of three parts: moment localization, classification, and saliency losses.  
Localization loss requires ground truth moment to measure the discrepancy with the predictions and can not be used in contrastive loss.
Thus, we design two-level contrastive losses based on two other losses: (1) proposal-level loss and (2) frame-level loss with a randomly sampled negative window. 

For the proposal-level contrastive loss ($\mathcal{L}_{p}$), 
proposals in the negative window are assigned with minimized scores compared to positive proposal~(i.e., the optimal proposal selected from the Hungarian algorithm) from the positive window, as follows,
\begin{equation}
\mathcal{L}_{p} =  - \sum_{p_i^{+} \in W^{+}} \log (s_i) - \sum_{p_i^{-} \in W^{-}} \log (1 - s_i)
\end{equation}
where $p_i^{+}$ is the positive proposal from the positive window $W^{+}$ and $p_i^{-}$ is each proposal from the negative window $W^{-}$, $s_i$ is the corresponding proposal score.

For the frame-level loss $\mathcal{L}_f$, we set the average saliency scores for frames located in the positive window is larger than the maximum saliency score of frames in the negative window over a margin $\delta$:
\begin{equation}
\small
     \mathcal{L}_{f}=\max(0, \delta + \max(S(W^{-},Q))-\textrm{mean}(S(W^{+},Q))
\end{equation}
where S() is its saliency scoring function.
So the overall contrastive loss is $\mathcal{L}_{con}=L_{p}+L_{f}$.

\paragraph{2D-TAN Base Model.} The original loss of 2D-TAN consists of the binary cross-entropy loss, which learns to align each proposal score with its scaled IoU value.
Thus, we assign each proposal in the negative window with minimized scores compared to the positive proposals~(i.e., the IoU with ground truth is large than 0.7) from the positive window, as follows,

\begin{equation}
\mathcal{L}_{p} =  - \sum_{p_i^{+} \in W^{+}} \log (s_i) - \sum_{p_i^{-} \in W^{-}} \log (1 - s_i)
\end{equation}
where $p_i^{+}$ is the positive proposal from the positive window $W^{+}$ and $p_i^{-}$ is each proposal from the negative window $W^{-}$, $s_i$ is the corresponding proposal score. And the overall contrastive loss is  $\mathcal{L}_{con}=L_{p}$.

\paragraph{Overall Loss.}
In total, our training loss ($\mathcal{L}$) consists of three parts: (1)  original training loss of the base model; (2)  contrastive loss to discriminate the negative window versus the positive window; (3) adapter loss (shown in Eq. \ref{equ: adapt}) to tune visual representations from general pre-training to the current downstream task:
\begin{equation}
    \mathcal{L} = L_{ori} + \lambda_{con} \times L_{con} + \lambda_{adapt} \times L_{adapt}
\end{equation}
where $\lambda_{con}$ and $\lambda_{adapt}$  are loss weight hyperparameters to control the loss value. During training, we set $\lambda_{con}$ to 1 and $\lambda_{adapt}$ to 0.2.


\section{More Qualitative Examples}
\label{sec: more-example}
We show four success cases~in Figure~\ref{fig:appendix_success} and four failure cases of our model \modelname~in Figure~\ref{fig:appendix_failure}.

